\documentclass[runningheads]{llncs}

\usepackage[T1]{fontenc}
\usepackage{graphicx}
\usepackage{tikz}
\usepackage{booktabs}
\usepackage{caption}
\usepackage{multirow}
\usepackage{amsmath, amssymb, amsfonts}
\usepackage[utf8]{inputenc}
\usepackage{placeins}
\usepackage{xcolor} 
\usepackage{float}
\usepackage{placeins} 

\usepackage[hyperindex,breaklinks]{hyperref}
\hypersetup{
    colorlinks=true,
    linkcolor=blue,   
    citecolor=blue,    
    filecolor=magenta, 
    urlcolor=blue     
}

\usepackage{tikz,xcolor,hyperref}

\definecolor{lime}{HTML}{A6CE39}
\DeclareRobustCommand{\orcidicon}{
	\begin{tikzpicture}
	\draw[lime, fill=lime] (0,0) 
	circle [radius=0.16] 
	node[white] {{\fontfamily{qag}\selectfont \tiny ID}};
	\draw[white, fill=white] (-0.0625,0.095) 
	circle [radius=0.007];
	\end{tikzpicture}
	\hspace{-2mm}
}

\foreach \x in {A, ..., Z}{\expandafter\xdef\csname orcid\x\endcsname{\noexpand\href{https://orcid.org/\csname orcidauthor\x\endcsname}
			{\noexpand\orcidicon}}
}
			
\usepackage{tikz}
\usepackage{xparse}   

\DeclareRobustCommand{\authorpic}[2][5mm]{%
  \tikz[baseline={([yshift=-.25ex]current bounding box.center)}]{%
    \clip (0,0) circle (#1);
    \pgfmathsetlengthmacro{\picside}{sqrt(2)*#1}%
    \node at (0,0) {\includegraphics[width=\picside,height=\picside,keepaspectratio]{#2}};
    \draw[line width=0.4pt, color=white] (0,0) circle (#1);
  }%
}

\NewDocumentCommand{\AuthorWithPic}{O{5.5mm} O{0.20em} m m}{%
  \texorpdfstring{\authorpic[#1]{#4}\kern #2}{}%
  #3%
}

\usepackage{xcolor}
\definecolor{linkpinkix}{HTML}{EA335A} 

\definecolor{linkpink}{HTML}{EA335A}

\newcommand{\shadedlink}[2]{%
  \tikz[baseline=(n.base)]\node[
    fill=linkpink,
    fill opacity=0.5,
    text opacity=1,
    rounded corners=.3ex,
    inner xsep=.35em,
    inner ysep=.15em
  ] (n) {\href{#1}{\textcolor{blue!70!black}{#2}}};%
}

\begin{document}
\title{Multimodal Sheaf-based Network for Glioblastoma Molecular Subtype Prediction}

\titlerunning{MMSN for Glioblastoma Molecular Subtype Prediction}  

\author{%
  \AuthorWithPic[6mm][0.18em]{Shekhnaz Idrissova}{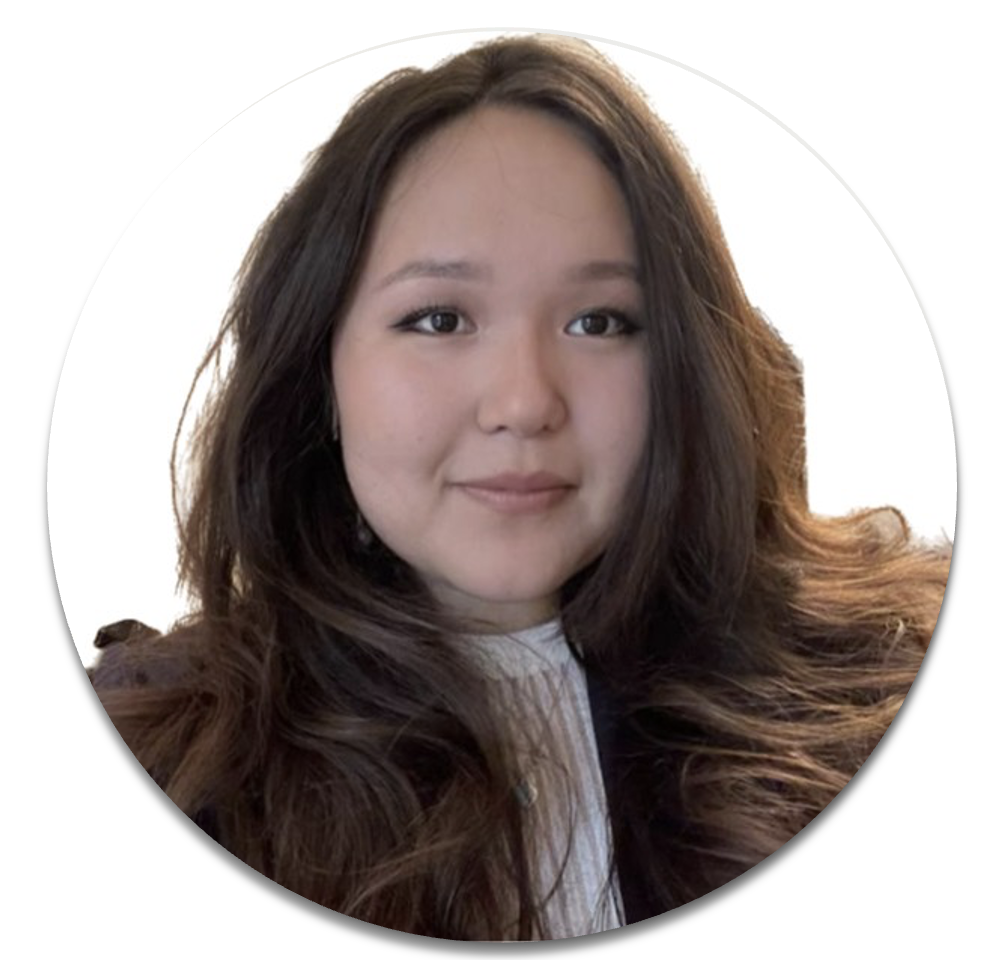} \and
  \AuthorWithPic[6mm][0.18em]{Islem Rekik\orcidA{}}{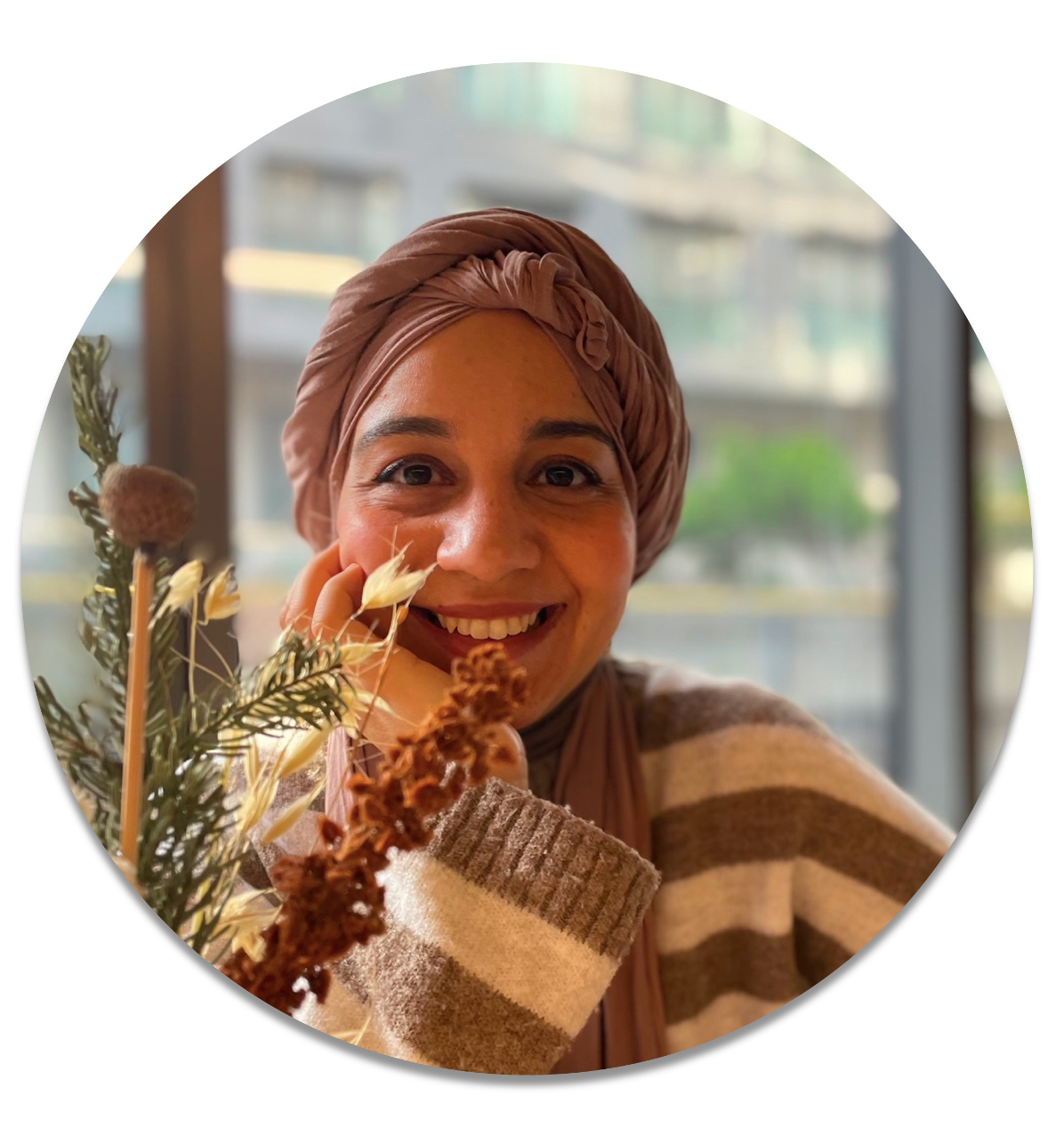}\thanks{Corresponding author: \email{i.rekik@imperial.ac.uk}, \url{http://basira-lab.com}, GitHub: \url{https://github.com/basiralab/MMSN}}%
}

\authorrunning{S Idrissova et al.}

\institute{BASIRA Lab, Imperial-X and Department of Computing, Imperial College London, United Kingdom}

\maketitle
%
\begin{abstract}
Glioblastoma is a highly invasive brain tumor with rapid progression rates. Recent studies have shown that glioblastoma molecular subtype classification serves as a significant biomarker for effective targeted therapy selection. However, this classification currently requires invasive tissue extraction for comprehensive histopathological analysis. Existing multimodal approaches combining MRI and histopathology images are limited and lack robust mechanisms for preserving shared structural information across modalities. In particular, graph-based models often fail to retain discriminative features within heterogeneous graphs, and structural reconstruction mechanisms for handling missing or incomplete modality data are largely underexplored. To address these limitations, we propose a novel sheaf-based framework for structure-aware and consistent fusion of MRI and histopathology data. Our model outperforms baseline methods and demonstrates robustness in incomplete or missing data scenarios, contributing to the development of virtual biopsy tools for rapid diagnostics. Our source code is available at \url{https://github.com/basiralab/MMSN/}.\footnote{This paper has been selected for a \textbf{Poster Presentation} at the AMAI MICCAI 2025 workshop. \shadedlink{https://youtu.be/3wQ_ahjieug}{[MMSN YouTube Video]}.}

\end{abstract}
\section{Introduction}

Glioblastoma is a grade IV astrocytoma, representing the most aggressive form of brain tumor, and is characterized by rapid proliferation of astrocytic-lineage cells. This highly aggressive malignancy carries a poor prognosis, with affected individuals typically surviving 12-18 months following diagnosis \cite{dewdney_signalling_2023}. The diversity of cell populations, driven by genetic alterations in oncogenic genes, has led to the identification of distinct glioblastoma molecular subtypes: classical, mesenchymal, neural, and proneural, with mixed subtypes also occurring \cite{drexler_molecular-based_2025}. Recent studies have identified molecular subtype classification as a significant biomarker for effective targeted therapy selection, though accurate classification requires thorough microscopic analysis of brain tumor tissue to detect specific cell populations present \cite{shikalov_targeted_2024}. 

Numerous studies have developed neural network models to predict wild-type \textit{IDH}, \textit{MGMT} methylation statuses from histopathology and MRI data separately, and survival predictions on the combination of two \cite{chakrabarty_mri-based_2023,li2023survival}. The use of both imaging modalities in clinical settings for a suitable diagnosis, where MRI provides macroscopic features such as tumor size, consistency, and location, while histopathology reveals microscopic characteristics that include cell granularity and infiltration patterns \cite{szopa2017diagnostic}. Both modalities contain inherent limitations: MRI suffers from signal processing artifacts, while histopathology is affected by staining variations and tissue preservation issues \cite{tseng2023histology,mikkelsen2021histological}. Despite these individual limitations, when combined, these complementary modalities provide a more comprehensive understanding of tumor characteristics. The spatial alignment between the two modalities is typically missing, and histopathology requires highly invasive procedures to obtain tissue samples. This work is motivated by the need to develop a comprehensive multimodal neural network capable of making molecular subtype predictions based on combined MRI and histopathology data, while also providing reconstruction capabilities for partially missing or completely absent modalities, following real-world clinical scenarios.

 The key contributions of our work are as follows:
\begin{enumerate}
\item We leverage the sheaf neural networks, to model higher-order relationships between different data spaces, to enable effective fusion of multiple imaging modalities.
\item We test the hypothesis that both modalities share an underlying topological structure of the tumor, which could explain their complementary diagnostic capabilities despite different scales and acquisition methods.
\item We provide comprehensive validation of our model across two critical scenarios: standard classification when both MRI and histopathology modalities are available, and when one modality is incomplete or missing.
\end{enumerate}

\section{Related work}
\textbf{Multimodal graph-based learning:} GNNs have been applied to multimodal fusion and representation learning, primarily using heterogeneous GNNs (HGNNs) for node classification and graph completion tasks rather than graph-level representation learning \cite{kim_heterogeneous_2023,akkus_multimodal_2023,peng_learning_2024}. However, these methods may suffer from over-smoothing, which can blur modality-specific features in heterogeneous graphs and hinder the preservation of discriminative information across modalities, as well as, requiring prior knowledge of inter-modal relationships. Cross-attention approaches have been developed to dynamically assign weights to modality importance by preprocessing modalities separately before fusing them in latent space, though these typically work with same-scale data (e.g., different MRI types) \cite{bi_cross-modal_2023,cai_graph_2023}. Co-attention mechanisms have also been used to find cross-modal similarity and cross-attention mappings through attention-based matching of uni-modal graph representations \cite{guo_cross-modal_2023,bi_cross-modal_2023}. The integration of histopathology and MRI images using the HMCAT architecture leverages Vision Transformers (ViT) and cross-attention mapping for enhanced multi-modal analysis \cite{li_survival_2023}. However, cross-attention-based methods may exhibit variable modality contributions, potentially resulting in imbalanced representational signals across input modalities. Therefore, balanced contribution and interactive fusion of multiple modalities are crucial to mimic comprehensive clinical expert reasoning by utilizing the distinct strengths of each modality.

\noindent\\
\textbf{Missing modality reconstruction} Missing modality reconstruction has gained significant interest, although most methods focus on training robust models rather than explicit reconstruction for downstream classification tasks \cite{wu_deep_2024}. Existing approaches include cross-attention fusion with available modalities and regularization techniques (e.g., HGMF), Gaussian mixture model compensator for missing graph node features integrated with GCNs, and attention-based or contrastive learning methods for handling complete modality absence \cite{nguyen_mi-cga_2025,taguchi_graph_2021,wu_multimodal_2023,zhao_graph_2025}. That said, studies specifically targeting reconstruction of graph node features from other modalities remain limited, with most work emphasizing model robustness in extreme missing-modality scenarios. Given the prevalence of noisy and missing medical data, our work addresses these challenges by enabling interactive modality fusion through sheaf neural networks and reconstructing missing features using learned sheaf operators \cite{hansen_sheaf_2020}.

\section{Methods}

\begin{figure}[!htbp]
  \centering
  \includegraphics[width=\textwidth]{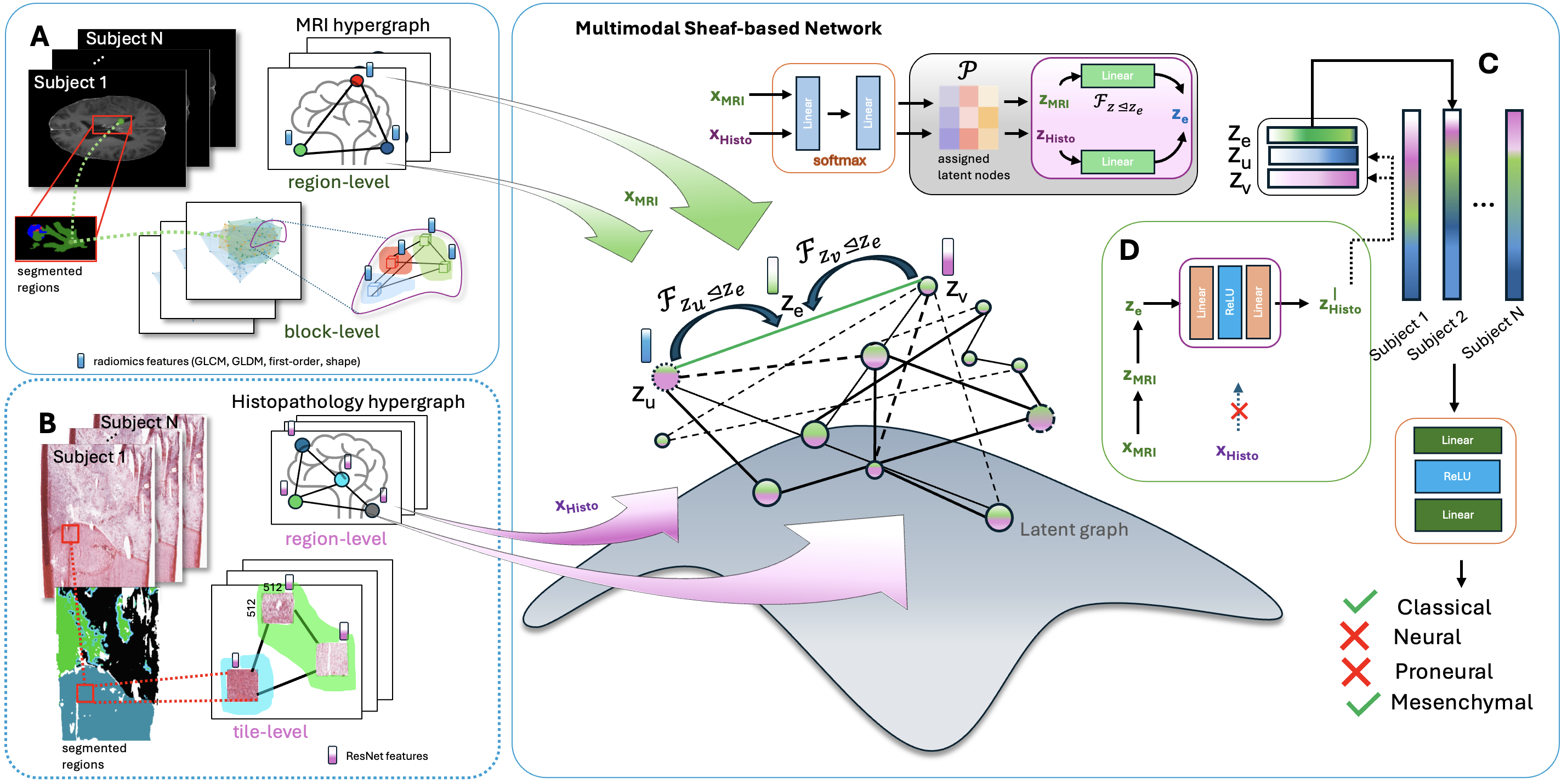}
  \caption{Proposed Multimodal Sheaf-based Network architecture (MMSN).  \textbf{A)} MRI-specific modality encoder.  \textbf{B)} Histopathology-specific modality encoder. \textbf{C)} Sheaf-based fusion on latent graph.  \textbf{D)} Missing modality reconstruction framework.}
  \label{fig:1}
\end{figure}

Recently, algebraic topology has been widely explored as an anchoring theorem to boost the design of novel interpretable deep learning models. While having robust applications, this paper primarily uses sheaf theory as an application for the fusion of multiple modalities, hypothesizing shared underlying topology between two input modalities. \textbf{Fig.~}\ref{fig:1} presents our proposed sheaf neural network architecture that addresses the challenge of integrating heterogeneous MRI and histopathology data for brain tumor analysis. The model leverages sheaf theory principles where local views from each modality naturally overlap within the global tumor context. Following image preprocessing, we construct region-level graphs capturing higher-order relationships in each modality (\textbf{Fig.~}\ref{fig:1}-\textbf{A}-\textbf{B}). Nodes are soft-assigned to a shared latent space, enabling multimodal fusion while preserving modality-specific information through restriction maps. The sheaf neural network facilitates controlled information diffusion, constraining each node's contribution to overlapping regions and maintaining the unique characteristics of each imaging modality (\textbf{Fig.~}\ref{fig:1}-\textbf{C}). The learned structure enables reconstruction of missing node features and derives final graph embeddings for tumor classification (\textbf{Fig.~}\ref{fig:1}-\textbf{D}). This approach provides a mathematically principled framework for multimodal medical data integration that preserves the complementary strengths of both MRI and histopathology imaging beyond simple fusion methods, while handling missing data.

 A \textbf{cellular sheaf} $(F, G)$ over an undirected graph $G = (V, E)$ attaches the algebraic data structure to each node $v \in V$ and edge $e \in E$ a vector space, called a \emph{stalk}, denoted by $F(v)$ and $F(e)$, respectively \cite{hansen_sheaf_2020}. 

 Additionally, every incident node-edge pair $v \trianglelefteq e$ (i.e., node $v$ is incident to edge $e$) is assigned a linear map between stalks, called a \emph{restriction map}, denoted as $F_{v \trianglelefteq e} : F(v) \to F(e)$.

 Given two nodes $v, u \in V$ connected by an edge $e = (v, u)$, a vector $x_v \in F(v)$ can be transported from $v$ to $u$ via the following sequence \cite{hansen_sheaf_2020}: $F(v) \xrightarrow{F_{v \trianglelefteq e}} F(e) \xrightarrow{F_{u \trianglelefteq e}^T} F(u).$

The sheaf Laplacian matrix is constructed by diagonal blocks, defined as 
\( L_{F_{vv}} = \sum_{v \trianglelefteq e} F^\top_{v \trianglelefteq e} F_{v \trianglelefteq e} \), 
and off-diagonal blocks as 
\( L_{F_{vu}} = - F^\top_{v \trianglelefteq e} F_{u \trianglelefteq e} \), 
where nodes \( u \) and \( v \) are connected by an edge \( e \).
The normalized sheaf Laplacian operator is then defined as 
\( \Delta_F = D^{-\frac{1}{2}} L_F D^{-\frac{1}{2}} \), 
and is used to compute the sheaf-generalized GCN layer: 
\( \mathrm{GCN}(X, A) := \sigma\left(D^{-\frac{1}{2}} A D^{-\frac{1}{2}} X W\right) = \sigma\left((I - \Delta_F) X W\right) \), 
as established in earlier works \cite{bodnar_neural_2023}.

\vspace{1em}

\textbf{Modality-specific encoders}
Given a set of input graphs \( \{G^{(p)}\} \) for each patient \( p \), we construct a corresponding higher-order hypergraph \( \mathcal{G}^{(p)} = (\mathcal{V}, \mathcal{E}) \). Each node \( v_i \in \mathcal{V} \) represents a graph element (e.g., a supernode or feature point), and each hyperedge \( e_j \in \mathcal{E} \) corresponds to a local region, defined based on spatial proximity or a shared label as illustrated in our main figure (\textbf{Fig.~}\ref{fig:1}).

 The incidence matrix \( \mathbf{H} \in \{0,1\}^{|\mathcal{V}| \times |\mathcal{E}|} \) is computed for the nodes belonging to the associated hyperedge set. A message-passing operation is performed from the node set to the hyperedge set. For each hyperedge \( e_j \), the aggregated embedding \( h_{e_j} \) is computed as:
\[
h_{e_j} = \text{AGGREGATE} \left( \{ h_{v_i} \mid H_{ij} = 1 \} \right)
\]
where \( h_{v_i} \) is the feature representation of node \( v_i \), and \( \text{AGGREGATE}(\cdot) \) is a permutation-invariant function (mean).

 A region-level graph is constructed where each node corresponds to a hyperedge \( e_j \) from the hypergraph (\textbf{Fig.~}\ref{fig:1}-\textbf{A}-\textbf{B}). Edges are defined based on the interconnectivity between local regions in the original input graph \( G^{(p)} \). Finally, a graph neural network (GNN) layer is applied to this region-level graph, updating the region-level node embeddings and forming the final modality-specific region-level graph. The higher-order construction captures both local region-level information and their inter-regional connectivity within each patient's graph.

\textbf{Cross-modality sheaf-based diffusion}
A latent graph \( \mathcal{G}_{\text{latent}} = (\mathcal{V}_\ell, \mathcal{E}_\ell) \) is constructed using an initial set of latent nodes \( \mathcal{V}_\ell \), each with learnable features \( h_{v_i}^{\ell} \in \mathbb{R}^d \). Edges \( \mathcal{E}_\ell \) are formed based on node proximity and feature similarity. Specifically, for any pair of latent nodes \( (v_i, v_j) \), an edge is established if a similarity function \( s(h_{v_i}^{\ell}, h_{v_j}^{\ell}) \) exceeds a threshold. The edge features \( e_{ij} \in \mathbb{R}^d \) are initialized upon edge creation. This latent graph is instantiated once and shared across all patient samples to serve as a global latent space.

 To align patient-specific data with the latent graph, a soft assignment mechanism is employed. Let \( x^{(m)} \in \mathbb{R}^{n_m \times d} \) denote the input node features for modality \( m \). A projection MLP followed by softmax defines the assignment probabilities to latent nodes (\textbf{Fig.~}\ref{fig:1}-\textbf{C}):
\[
P^{(m)} = \text{softmax} \left( \text{MLP}(x^{(m)}) \right) \in \mathbb{R}^{n_m \times |\mathcal{V}_\ell|}
\]
These soft assignment scores define the probability that a node from modality \( m \) is associated with each latent node. The projected features in the latent space are computed as:
\[
\hat{X}^{(m)} = P^{(m)\top} x^{(m)}
\]

 Following the soft node assignment, we define a sheaf \( \mathcal{F} \) over the latent graph \( \mathcal{G}_{\text{latent}} \). Each node \( v \in \mathcal{V}_\ell \) is associated with a stalk \( \mathcal{F}(v) = \mathbb{R}^d \), and each edge \( e = (u,v) \in \mathcal{E}_\ell \) is associated with a stalk \( \mathcal{F}(e) = \mathbb{R}^d \) representing the overlapping region. For every edge \( e = (u,v) \), we define learnable linear restriction maps:
\[
\rho_{e,u} : \mathcal{F}(u) \to \mathcal{F}(e), \quad \rho_{e,v} : \mathcal{F}(v) \to \mathcal{F}(e)
\]
These maps project node-level features to the shared space defined on the edge.

 A sheaf-based GCN layer is then applied to propagate information through the latent graph using these restriction maps. Finally, a patient-specific latent representation is formed by concatenating the sum of node-level latent features and the sum of edge-level features:
\[
h_{\text{patient}} = \text{CONCAT} \left( \sum_{v \in \mathcal{V}_\ell} h_v^\ell, \quad \sum_{e \in \mathcal{E}_\ell} e \right)
\]

 This representation \( h_{\text{patient}}\) is used as input to downstream classification tasks.

\textbf{Modality reconstruction}
The learned restriction maps \( \rho_{e,v} \), previously defined as part of the sheaf neural network, serve as structural anchors during reconstruction. 
These projected features \( \hat{X}^{(\text{obs})} \) are used to update node-level features \( h_v^\ell \) in the latent graph.

 For each latent node \( v \in \mathcal{V}_\ell \), the associated edge stalk features are computed using the sheaf restriction maps $e_v = \sum_{e \in \mathcal{E}(v)} \rho_{e,v}(h_v^\ell)$. These edge-aggregated latent features \( e_v \in \mathbb{R}^{d} \) are passed through a reconstruction MLP to obtain the reconstructed node features for the missing modality using:
\begin{equation}
    \tilde{x}^{(\text{miss})}_v = \text{MLP}_{\text{recon}}(e_v)
\end{equation} 

The reconstruction loss is defined as a mean squared error (MSE) between the reconstructed node features and the original (masked) ones:
\begin{equation}
\mathcal{L}_{\text{recon}} = \sum_{v \in \mathcal{V}_{\text{miss}}} \left\| x_v^{(\text{miss})} - \tilde{x}_v^{(\text{miss})} \right\|_2^2
\end{equation}

The final patient-specific graph representation includes the reconstructed features of the missing modality. 
This representation is then used for downstream tasks such as classification. The consistency loss acts as a regularizer by enforcing agreement between node features projected onto the shared edge space through the restriction maps. The total training loss combines classification, reconstruction, and consistency objectives:

\begin{equation}
\mathcal{L}_{\text{total}} = \lambda_1 \mathcal{L}_{\text{classification}} + \lambda_2 \mathcal{L}_{\text{recon}} + \lambda_3 \mathcal{L}_{\text{consistency}}
\end{equation}

\section{Experiments}

\textbf{Dataset} The dataset is from the open-source Ivy GAP study, which provides an extensive range of glioblastoma data from 41 patients. The dataset is open-source and provides a comprehensive transcriptomic atlas along with molecular subtype annotations for each patient. While ethnicity data is not reported, the age distribution is concentrated between 50 and 69 years, with a relatively balanced representation of male and female patients. The MRI modalities include T1, T2, FLAIR, and T1-weighted, with segmented regions \cite{bakas_university_2022,puchalski_anatomic_2018}. We use adjacent H\&E-stained histopathology whole-slide images (WSI) from the Cancer Stem Cells ISH Survey study. Given the limited availability of paired H\&E histopathology and MRI data for glioblastoma, we identified 30 patients with complete datasets meeting our quality criteria. 

\textbf{Preprocessing} The MRI 3D graph is constructed by extracting 3D volume blocks from the original input data. These blocks come from non-overlapping regions defined by annotation masks, with block extraction proportional to the region distribution in the original MRI data. Each block forms a node in the 3D graph, and inter-regional edges are defined based on k-nearest neighbors (kNN) and intra-regional edges based on local labels. Node features are extracted using radiomic features including first-order statistics, GLCM, GLDM features from PyRadiomics package \cite{van_griethuysen_computational_2017}. For histopathology, 512×512 sized tiles are extracted from each non-overlapping region in annotated masks. A pretrained ResNet architecture is used to derive tile features \cite{he_deep_2015}. Each tile forms a node in the histopathology graph, and edges are formed based on kNN and node labels. The MRI and histopathology graphs are constructed per patient and serve as input to the model.

\textbf{Benchmarks} For both modality present case, the benchmark models included: \textbf{(1)} Multimodal Hypergraph Attention Network (MHAN) using cross-attention mapping for hypergraphs constructed from the original modality-specific graphs \cite{kim_hypergraph_2020};  \textbf{(2)} Fusion-GCN modality specific GCN encoders and late fusion with aggregation of two graph embeddings \cite{duhme_fusion-gcn_2021};  and \textbf{(3)} MGCL using multimodal graph contrastive learning  \cite{liu_multimodal_2023}. For the missing or incomplete modality scenario,  \textbf{(4)} Adversarial Attention-based Variational Graph Autoencoder (AAVGA) was used based on  \cite{weng_adversarial_2020}, where the cross-attention layer within the latent space is utilized. 

All benchmark models were trained and tested on the same dataset, with 3-fold cross validation. The models were tested using Adam optimizer \cite{kingma_adam_2017} and scheduler. The early stopping was initiated for all models. All models were trained for 100 epochs. The evaluation metrics included accuracy, sensitivity, specificity, macro- and micro-F1 scores.

\section{Results}
\begin{figure}[!htb]
    \centering
    \includegraphics[width=\textwidth]{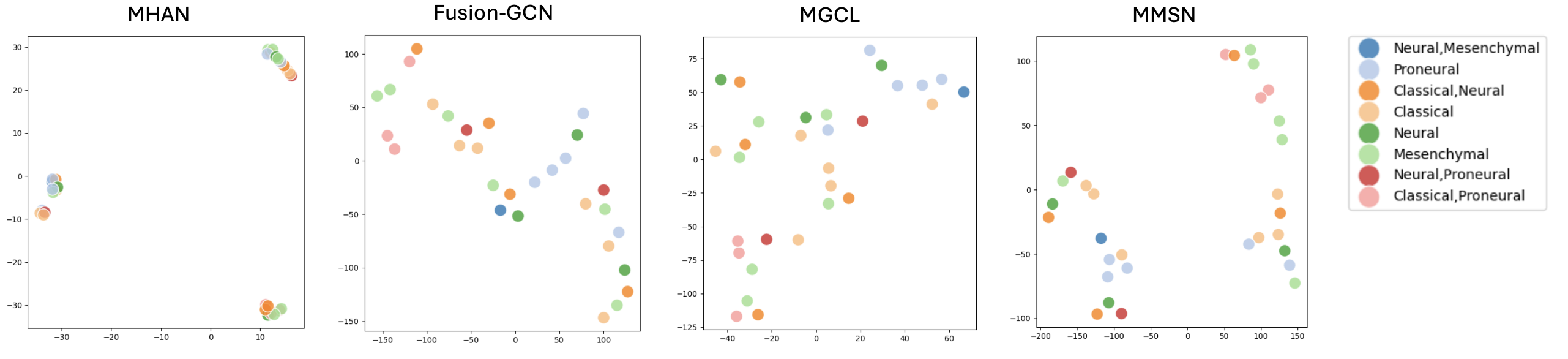}
    \caption{t-SNE plots for baseline and proposed model results.}
    \label{fig:2}
\end{figure}

As shown in \textbf{Table}~\ref{tab1}, our model outperforms the benchmarks in the multi-label classification task, which involves predicting the presence of molecular subtype characteristics—namely classical, neural, proneural, and mesenchymal. \textbf{Fig.}~\ref{fig:2} illustrates class distribution patterns via t-SNE visualization. While the MHAN model shows more distinct visual clustering, notable class overlap remains evident across all methods. The MMSN model demonstrates reasonable class separation while maintaining representation of the inherent feature complexity within each molecular subtype. The limited dataset size should be considered when interpreting these results, as most baseline models were originally developed and validated on larger datasets. Results demonstrate that our model's structural inductive biases enable effective performance despite limited data availability, achieving the highest accuracy (61.46\%) and macro-F1 score (50.82\%) among all compared methods, suggesting potential benefits from leveraging topological relationships between modalities. In Table~\ref{tab1}, the proposed model exhibits higher sensitivity than specificity, in contrast to the baselines. This may reflect the natural limitations of a small dataset, as well as the model's tendency to favor detecting multiple subtype signatures. Such behavior is consistent with the biological reality of glioblastoma’s intratumoral heterogeneity, where distinct molecular subtypes can coexist within a single tumor.

For both conditions, where the modality is missing or not, before the evaluation, both benchmark and proposed models were trained with randomly dropped histopathology data during training. \textbf{Table}~\ref{tab2} shows our MMSN model consistently outperforms the AAVGA benchmark across various dropout rates, though with some performance variability. These results suggest that for the glioblastoma diagnostic imaging data, the geometric structural approaches may offer advantages over the tested variational graph autoencoder method, particularly given the importance of anatomical structure in medical imaging analysis. 
\vspace{-10pt}
\begin{table}[!htbp]
\centering
\caption{Evaluation metrics results from proposed model and baselines, with complete set of both modalities.}
\label{tab1}
\begin{tabular}{lccccc}
\toprule
Model & Accuracy & Sensitivity & Specificity & Macro-F1 & Micro-F1 \\
\midrule
MHAN & 49.72 & 41.67 & 58.33 & 19.37 & 31.51 \\
Fusion-GCN & 49.72 & 49.93 & 54.82 & 31.76 & 36.75 \\
MGCL & 51.04 & 50.56 & 56.21 & 35.86 & 38.95 \\
\textbf{MMSN (ours)} & \textbf{61.46} & \textbf{66.39} & \textbf{57.68} & \textbf{50.82} & \textbf{54.1} \\
\bottomrule
\end{tabular}
\end{table}
\vspace{-10pt}
\vspace{-10pt}
\begin{table}[!htbp]
\centering
\caption{Evaluation metrics for AAVGA and MMSN across different dropout rates ($p$) during training.}
\label{tab2}
\begin{tabular}{clccccc}
\toprule
$p$ & Model & Accuracy & Sensitivity & Specificity & macro-F1 & micro-F1 \\
\midrule
\multirow{2}{*}{0.00} & AAVGA & 42.36 & \textbf{58.33} & 41.67 & 26.37 & 35.94 \\
                      & \textbf{MMSN} & \textbf{60.44} & 43.89 & \textbf{65.5} & \textbf{33.0} & \textbf{44.5} \\
\multirow{2}{*}{0.25} & AAVGA & 41.67 & \textbf{56.11} & 37.85 & \textbf{28.57} & 35.7 \\
                      & \textbf{MMSN} & \textbf{49.32} & 52.08 & \textbf{47.62} & 28.04 & \textbf{41} \\
\multirow{2}{*}{0.50} & AAVGA & 49.77 & \textbf{55.28} & 48.02 & 31.69 & \textbf{42.42} \\
                      & \textbf{MMSN} & \textbf{56.22} & 42.22 & \textbf{64} & \textbf{33.94} & 36.76 \\
\multirow{2}{*}{0.75} & AAVGA & \textbf{48.15} & 48.47 & \textbf{52.11} & 32.29 & 35.92 \\
                      & \textbf{MMSN} & 47.36 & \textbf{50.42} & 43.33 & \textbf{33.68} & \textbf{40.4} \\
\multirow{2}{*}{1.00} & AAVGA & 51.85 & 55.28 & \textbf{52} & 36.68 & 44.44 \\
                      & \textbf{MMSN} & \textbf{54.74} & \textbf{63.19} & 45.48 & \textbf{40.25} & \textbf{49.11} \\
\bottomrule
\end{tabular}
\end{table}
\vspace{-10pt}
\section{Conclusion}
We introduce a novel geometric sheaf-based model for multimodal graph representation learning, hypothesizing that sheaf structures can preserve the intrinsic topological organization of glioblastoma by coordinating local feature representations from MRI and histopathology data. While the proposed sheaf-based architecture outperforms existing multimodal graph fusion baselines, its computational complexity constrains missing modality imputation performance on small datasets. Future research will aim to improve model efficiency and generalizability using larger datasets, as the current study was constrained by the limited size of the available open-source dataset containing paired histopathology and MRI modalities. With growing interest in this area, subsequent work will focus on incorporating additional non-imaging modalities and expanding topology-consistent fusion frameworks to better handle diverse multimodal scenarios.

\textbf{Prospect of application} This work introduces a structure-aware multimodal framework that improves robustness to missing data, advancing the development of virtual biopsy tools that could reduce reliance on invasive brain tumor tissue extraction procedures. By leveraging accessible radiomic tools such as MRI, and reconstructing the missing microscopic features of histopathology, this approach could contribute to faster and more efficient glioblastoma diagnosis, reducing the critical waiting time between initial presentation and treatment initiation—a factor crucial for patient survival outcomes.

\textbf{Disclosure of interests} The authors have no competing interests to declare that are relevant to the content of this article.

%
%
%
\bibliographystyle{splncs}
\bibliography{MICCAI_AMAI_2025}

\end{document}